\pdfoutput=1

\documentclass[11pt]{article}

\usepackage[]{EACL2023}

\usepackage{times}
\usepackage{latexsym}

\usepackage[T1]{fontenc}

\usepackage[utf8]{inputenc}

\usepackage{microtype}

\usepackage{inconsolata}

\usepackage{multirow}
\usepackage{graphicx}
\usepackage{booktabs}
\usepackage{amssymb}
\usepackage{amsmath}
\usepackage{pifont}
\usepackage{cleveref}
\usepackage{enumitem}
\crefformat{section}{\S#2#1#3} 
\crefformat{subsection}{\S#2#1#3}

\title{Plausible May Not Be Faithful: Probing Object Hallucination in Vision-Language Pre-training}
\author{Wenliang Dai$^\dagger$\quad~Zihan Liu\thanks{$^*$ Work done during PhD at HKUST.} $^{\text{ }\diamond}$\quad~Ziwei Ji$^\dagger$\quad~Dan Su$^\dagger$\quad~Pascale Fung$^\dagger$ \\
$^\dagger$The Hong Kong University of Science and Technology\\
$^\diamond$NVIDIA\\
\texttt{wenliang.dai@connect.ust.hk}, \texttt{pascale@ece.ust.hk}}

\begin{document}
\maketitle
\begin{abstract}
Large-scale vision-language pre-trained (VLP) models are prone to hallucinate non-existent visual objects when generating text based on visual information. In this paper, we systematically study the object hallucination problem from three aspects. First, we examine recent state-of-the-art VLP models, showing that they still hallucinate frequently, and models achieving better scores on standard metrics (e.g., CIDEr) could be more unfaithful. Second, we investigate how different types of image encoding in VLP influence hallucination, including region-based, grid-based, and patch-based. Surprisingly, we find that patch-based features perform the best and smaller patch resolution yields a non-trivial reduction in object hallucination. Third, we decouple various VLP objectives and demonstrate that token-level image-text alignment and controlled generation are crucial to reducing hallucination. Based on that, we propose a simple yet effective VLP loss named ObjMLM to further mitigate object hallucination. Results show that it reduces object hallucination by up to 17.4\% when tested on two benchmarks (COCO Caption for in-domain and NoCaps for out-of-domain evaluation). 
\end{abstract}

\section{Introduction} \label{sec:intro}

Thanks to the advancement of pre-trained large Language Models (LLMs) and Vision-Language Pre-training (VLP) methods, models are able to achieve surprisingly good performance in vision-conditioned text generation, e.g., image captioning.
However, LLMs are found to generate unfaithful or nonsensical texts given the source input~\citep{ji2022survey}, which is called hallucination. 
The hallucination problem is also inherited by VLP models~\citep{Alayrac2022FlamingoAV}, as they are still language model that can understand visual signals.
VLP models often generate fluent and seem appropriate sentences if we only see the text, but wrong when taking the visual input into consideration. 
One major type of hallucination in VLP is known as object hallucination~\citep{rohrbach2018object}, where models generate texts containing non-existent or inaccurate objects from the input images. 
Object hallucination in VLP models essentially limits their performance and raises safety concerns for industrial applications.
For example, in biomedical image captioning~\citep{biomedical_ic}, object hallucination reduces the accuracy of diagnosis and may lead to severe consequences for the patient. 
Despite the limitations and risks caused by object hallucination, 
this problem has not been studied in contemporary VLP works yet.

To narrow down the aforementioned research gap, in this paper,
we systematically investigate four fundamental research questions about object hallucination: 
1) how much do modern VLP models hallucinate?
2) how do different forms of image encoding in VLP affect object hallucination?
3) what are the effects of common VLP objectives on object hallucination? and
4) how to mitigate object hallucination in VLP models?


For our first question, we examine recent state-of-the-art VLP models on the image captioning task. To evaluate object hallucination, we adopt and expand the CHAIR (Caption Hallucination Assessment with Image Relevance) metric proposed by \citet{rohrbach2018object}. Results show that these models still hallucinate frequently with $\sim$10\% of the generated sentences containing at least one hallucinated object.
This problem becomes much severer when generating sentences given out-of-domain images. 
Furthermore, we discover that the widely used optimization method SCST~\citep{Rennie2017SelfCriticalST} could lead to more hallucination, even if it improves standard metrics like CIDEr~\citep{Vedantam2015CIDErCI}. While \citet{rohrbach2018object} observe a similar finding, we evaluate with a more diverse model pool, showing that large-scale VLP could not resolve this problem.


For our second question, to investigate how different types of image encoding in VLP influence hallucination, we ablate three commonly used ones, including region-based, grid-based, and patch-based~\citep{Kim2021ViLTVT}. Surprisingly, we find that patch-based features perform the best and smaller patch resolution yields a non-trivial reduction in object hallucination. 

Thirdly, we analyze the effects of commonly adopted vision-language pre-training objectives on object hallucination. Specifically, we decouple and combine the image-text contrastive (ITC) loss, the image-text matching (ITM) loss with and without hard negatives, and the image-conditioned language modeling (ICLM) loss. Counter-intuitively, although ITC and ITM help to bring apart dissimilar images and texts, results show that they do not contribute much to alleviating object hallucination. The generative ICLM loss is the main influential factor of object hallucination and different pre-training datasets lead to distinctive model behaviors. More detailed analysis is described in Section~\ref{sec:analysis_vlp_objectives}.

Finally, 
we propose a simple yet effective new vision-language pre-training loss, namely object-masked language modeling ({ObjMLM}), to further mitigate object hallucination by enhancing the alignment and restriction between text tokens and visual objects during generation. Code and evaluation setups are released: \url{https://github.com/wenliangdai/VLP-Object-Hallucination}.

Overall, our contributions are three-fold:
\vspace{-0.5em}
\begin{itemize}[itemsep=0.1em]
    \item This is the first paper that systematically studies state-of-the-art VLP models on the object hallucination problem, proving that it is still far from resolved and previous methods that improve standard metrics may reflect in worse hallucination.
    \item We thoroughly investigate the influence of different VLP losses and image encoding methods on object hallucination. Our findings could be valuable for future work to build more responsible VLP systems.
    \item We present a new pre-training objective \text{ObjMLM} to mitigate object hallucination. Experimental results show that it reduces object hallucination by 17.4\% without introducing extra training data.
\end{itemize}

\section{Related Work} \label{sec:related_work}

\subsection{Hallucination in Deep Learning}
Generally, the term \textit{hallucination} denotes the appearance of undesirable output that is unfaithful to the conditional input~\cite{maynez2020faithfulness}, even though it may appear to be fluent or reasonable. 
In the multimodal field, the hallucination phenomenon refers to the prediction of non-existent or incorrect objects (e.g., in object detection or image captioning) and is called \textit{object hallucination}~\citep{rohrbach2018object,biten2022let}.
Despite the success of large pre-trained models, they still suffer the hallucination problem, which degrades the performance and largely hinders practical applications~\cite{ji2022survey}.

Many works have been proposed to mitigate hallucination in recent years.
\citet{nie2019simple} applied data refinement with self-training to improve the equivalence between the input and the paired text in the data-to-text generation task.
\citet{Zhang_Shi_Tang_Xiao_Yu_Zhuang_2021} and \citet{10.1145/3394171.3413746} proposes scene graph learning methods to ground the process of visual captioning to reduce hallucination.
\citet{ma2020learning} reconstruct generated sentences from localized image regions.
\citet{xiao2021hallucination} proposed the uncertainty-aware beam search as an add-on technique to the original beam search, in both image captioning and data-to-text generation. 
To reduce hallucination in dialog systems, \citet{shuster2021retrieval} introduced knowledge augmentation and \citet{dziri2021neural} presented a post-processing method to refine generated outputs. \citet{su2022read} augment models with answer-related information predicted by a machine reading comprehension module to reduce hallucination in the generative question answering task.

\subsection{Vision-Language Pre-training} \label{sec:related_work_vlp_ic}
The research on vision-language pre-training (VLP) has progressed vastly in recent years.
Due to the demand for large-scale data, most VLP methods use self-supervised pre-training objectives to utilize image-text pairs crawled from the web.
In the beginning, BERT~\citep{Devlin2019BERTPO}-style VLP models~\citep{Lu2019ViLBERTPT,Tan2019LXMERTLC,Li2019VisualBERTAS,Chen2020UNITERUI,Yu_Tang_Yin_Sun_Tian_Wu_Wang_2021,Shen2021HowMC} are trained to perform multimodal understanding tasks, using objectives like image-text matching and masked language modeling. 
Later, encoder-decoder architectures are introduced to additionally handle multimodal generation tasks with a causal language modeling loss~\citep{Li2021UNIMOTU,yu-etal-2021-vision,Lin2021M6AC,Cho2021UnifyingVT,Ding2021CogViewMT,Li2022BLIPBL,Wang2022UnifyingAT}.
Another line of research uses a dual-stream architecture~\citep{Radford2021LearningTV,Jia2021ScalingUV,Zhai2021LiTZT,Yao2021FILIPFI} with separate image and text encoders aligned together through an image-text contrastive loss.
They improve the performance of various multimodal downstream tasks by a large step. 

\citet{Alayrac2022FlamingoAV} show that fatal object hallucination can happen naturally or be provoked by the adversarial prompting in modern VLP models. However, in previous works, how different VLP strategies influence the faithfulness of generated text given images has not been studied. Moreover, the effects of using different types of image encoding are also unclear, including region-based~\citep{Li2020OscarOA,Zhang2021VinVLRV,Hu2021ScalingUV}, grid-based~\citep{Wang2021SimVLMSV}, and patch-based~\citep{Kim2021ViLTVT,Li2021AlignBF}.


\section{Evaluation Setup}

In this section, we first introduce the CHAIR evaluation metric for automatic evaluation in Section~\ref{sec:eval_metric}. Then, in Section~\ref{sec:eval_datasets}, we describe two datasets that are used for testing and explain how to calculate CHAIR scores under such settings.

\subsection{Evaluation Metric} \label{sec:eval_metric}
To automatically measure object hallucination, we adopt the CHAIR (Caption Hallucination Assessment with Image Relevance) metric proposed by~\citet{rohrbach2018object}. 
CHAIR calculates what proportion of generated object words are not in the image (i.e., hallucinated) according to the ground truth. 
CHAIR has two variants: CHAIR$_i$ (instance-level) and CHAIR$_s$ (sentence-level), which are formulated as follows:
\begin{equation*}
    \textrm{CHAIR}_i = \frac{\textrm{\# \{hallucinated objects\}}}{\textrm{\# \{all objects in prediction\}}},
\end{equation*}
\begin{equation*}
    \textrm{CHAIR}_s = \frac{\textrm{\# \{hallucinated sentences\}}}{\textrm{\# \{all sentences\}}}.
\end{equation*}
As formulated, CHAIR$_i$ represents the proportion of hallucinated objects over all golden objects in all data samples. It can be seen as the probability of a generated object to be a hallucination.
On the other hand, CHAIR$_s$ measures the proportion of generated sentences that contain at least one hallucinated object. Therefore, to calculate CHAIR$_i$ and CHAIR$_s$, we need a pre-defined list of golden object categories to recognize objects in the text. We illustrate dataset-specific calculation details in Section~\ref{sec:eval_datasets}.


\begin{table*}[t]
\centering
\resizebox{\textwidth}{!}{
\begin{tabular}{l|c|c|cccccc|cccc}
\toprule
\multirow{3}{*}{Model} & \multirow{3}{*}{\begin{tabular}[c]{@{}c@{}}CIDEr\\ Optim\\(SCST)\end{tabular}}  & \multicolumn{1}{c|}{\multirow{3}{*}{\begin{tabular}[c]{@{}c@{}}\# Pretrain\\Image-Text\\Pairs\end{tabular}}} & \multicolumn{6}{c|}{\multirow{2}{*}{COCO Caption Karpathy Test}} & \multicolumn{4}{c}{NoCaps Validation} \\
 &  & \multicolumn{1}{c|}{} & \multicolumn{6}{c|}{}  & \multicolumn{4}{c}{Out-of-domain} \\ 
 &  & \multicolumn{1}{c|}{} & B@4$\uparrow$ & C$\uparrow$ & M$\uparrow$ & S$\uparrow$ & {CH$_i$}$\downarrow$ & {CH$_s$}$\downarrow$  & C$\uparrow$ & S$\uparrow$ & {CH$_i$}$\downarrow$ & {CH$_s$}$\downarrow$ \\ \midrule \midrule
OSCAR \small{\textit{Base}}$^*$ & \ding{55} & 6.5M & 34.4 & 117.6 & 29.1 & 21.9 & 7.1  & 13.0  & - & - & - & - \\
OSCAR \small{\textit{Base}}$^*$ & \ding{51} &6.5M &39.6 & 134.2 & 29.8 & 23.5 & 7.2  & 13.5  & - & - & - & - \\
VinVL \small{\textit{Base}} & \ding{55} & 6.5M&38.2 & 129.3 & 30.3 & 23.6 & 5.3  & 10.0  & 83.1 & 10.8 & 12.1 & 21.2 \\
VinVL \small{\textit{Base}} & \ding{51} &6.5M &40.9 & 140.4 & 30.9 & 25.1 & 5.7  & 10.9  & 87.5 & 11.7 & 17.4 & 32.1 \\
VinVL \small{\textit{Large}} & \ding{55} &6.5M& 38.5 & 130.8 & 30.4 & 23.4 & 5.5 & 10.5  & - & - & - & - \\
VinVL \small{\textit{Large}} & \ding{51}  &6.5M&41.0 & 140.9 & 31.1 & 25.2 & 5.6 & 10.6 & - & - & - & - \\
BLIP \small{\textit{Base}} & \ding{55} &129M&39.7 & 133.3 & 31.0 & 23.8 & 4.9 & 8.9 & 112.1 & 14.2 & 6.6 & 10.5 \\
BLIP \small{\textit{Large}} & \ding{55}  &129M&40.4 & 136.7 & 31.1 & 24.3 & 4.7 & 8.8 & 115.3 & 14.4 & 6.4 & 10.5 \\
OFA \small{\textit{Large}} & \ding{55} & 21M$^\dagger$ &41.7 & 140.5 & 31.2 & 24.2 & 4.7 & 8.9 & 103.2 & 13.3 & 6.4 & 10.2 \\
OFA \small{\textit{Large}} & \ding{51} & 21M$^\dagger$ &43.8 & 149.5 & 31.8 & 25.9 & 4.2 & 8.1 & 113.1 & 15.2 & 7.1 & 12.4 \\  \bottomrule
\end{tabular}
}
\caption{Image captioning results of recent state-of-the-art VLP models~\citep{Li2020OscarOA,Zhang2021VinVLRV,Li2022BLIPBL,Wang2022UnifyingAT} on the COCO Caption Karpathy test set and NoCaps validation set. Here, B@4, C, M, S, and CH denote BLEU-4, CIDEr, METEOR, SPICE, and CHAIR, respectively. CIDEr Optim indicates whether the SCST CIDEr optimization is used or not. All results are generated by using their officially provided checkpoints and hyper-parameters, * means the model is finetuned by us as the provided one is broken. $\dagger$ denotes the model also uses unimodal data besides image-text pairs.}
\label{tab:sota_table}
\end{table*}

\subsection{Evaluation Datasets}  \label{sec:eval_datasets}
To evaluate models' performance on object hallucination with CHAIR, we adopt two widely used benchmarks: Microsoft COCO Caption~\citep{Lin2014MicrosoftCC} and NoCaps~\citep{Agrawal2019nocapsNO}. For all models, the COCO Caption training set is used for the finetuning of the image captioning task, and COCO Caption test set and NoCaps valid set are used for in-domain and out-of-domain evaluation, respectively. In the following, we introduce statistics of each dataset and how to calculate CHAIR on them.

\subsubsection{COCO Caption} The COCO Caption~\citep{Lin2014MicrosoftCC} is a large-scale and widely used dataset for the training and evaluation of the image captioning task. We use the Karpathy split~\citep{Karpathy2017DeepVA}, in which 82K, 5K, and 5K images are in the train, validation, and test sets, respectively. Each image is annotated with at least five ground truth captions. 

To calculate CHAIR scores on this dataset, we follow the setting proposed in \citet{rohrbach2018object}. In practice, we first tokenize each sentence and then singularize each word. Then, we use a list of synonyms from~\citet{Lu2018NeuralBT} to map fine-grained objects to the pre-defined 80 coarse-grained MSCOCO object categories (e.g., mapping ``puppy'', ``chihuahua'', ``poodle'' to the ``dog'' category). The purpose of doing this mapping is to ensure that we do not detect hallucinated objects by mistake. For example, when the ground-truth caption only has the ``puppy'' object, the CHAIR metrics will undesirably consider the ``dog'' object generated by models as a hallucinated object if we do not perform the mapping. 


\subsubsection{NoCaps} The NoCaps~\citep{Agrawal2019nocapsNO} dataset aims to evaluate models trained on the training set of COCO Caption to examine how well they generalize to a much larger variety of visual concepts, i.e., unseen object categories. There are 4,500 images in the validation set and 10,600 images in the test set. Images are taken from the Open Images V4~\citep{Kuznetsova2020TheOI} dataset, which contains 600 object classes. Due to the unavailability of ground truth captions of the test set, we use the valid set of NoCaps. 

To calculate CHAIR scores on NoCaps, 
we setup a similar setting as used in COCO Caption. 
Specifically, we map the fine-grained classes defined in NoCaps to coarse-grained categories based on the hierarchical object relationship\footnote{\url{https://github.com/nocaps-org/image-feature-extractors/blob/master/data/oi_categories.json}} to improve the effectiveness of CHAIR metrics. We only add two types of object categories to our final object list: 1) super-categories that have sub-categories, and 2) object categories that have neither super-category nor sub-categories. Eventually, we construct a list of 139 coarse-grained object categories from the 600 classes.

\section{Object Hallucination in VLP Models} \label{sec:sota_models}
Benefitting from the vast advancement of various VLP methods, the performance of image captioning has been improved a lot by following a pretrain-then-finetune schema. Generally, the performance is measured by metrics like CIDEr~\citep{Vedantam2015CIDErCI}, SPICE~\citep{Anderson2016SPICESP}, METEOR~\citep{Banerjee2005METEORAA}, and BLEU~\citep{Papineni2002BleuAM}, which consider the semantic and syntactic similarity or n-gram-based fluency between the model generated and ground truth captions. However, the faithfulness of captions generated by VLP models is neglected.

In this section, we provide a thorough analysis of recent VLP models to investigate how much they hallucinate when generating text conditioned on visual information. The results are shown in Table~\ref{tab:sota_table}. Models are finetuned on the COCO Caption training set and evaluated on both the COCO Caption test set and the NoCaps valid set. 

\begin{figure}[t]
    \centering
    \includegraphics[width=0.96\linewidth]{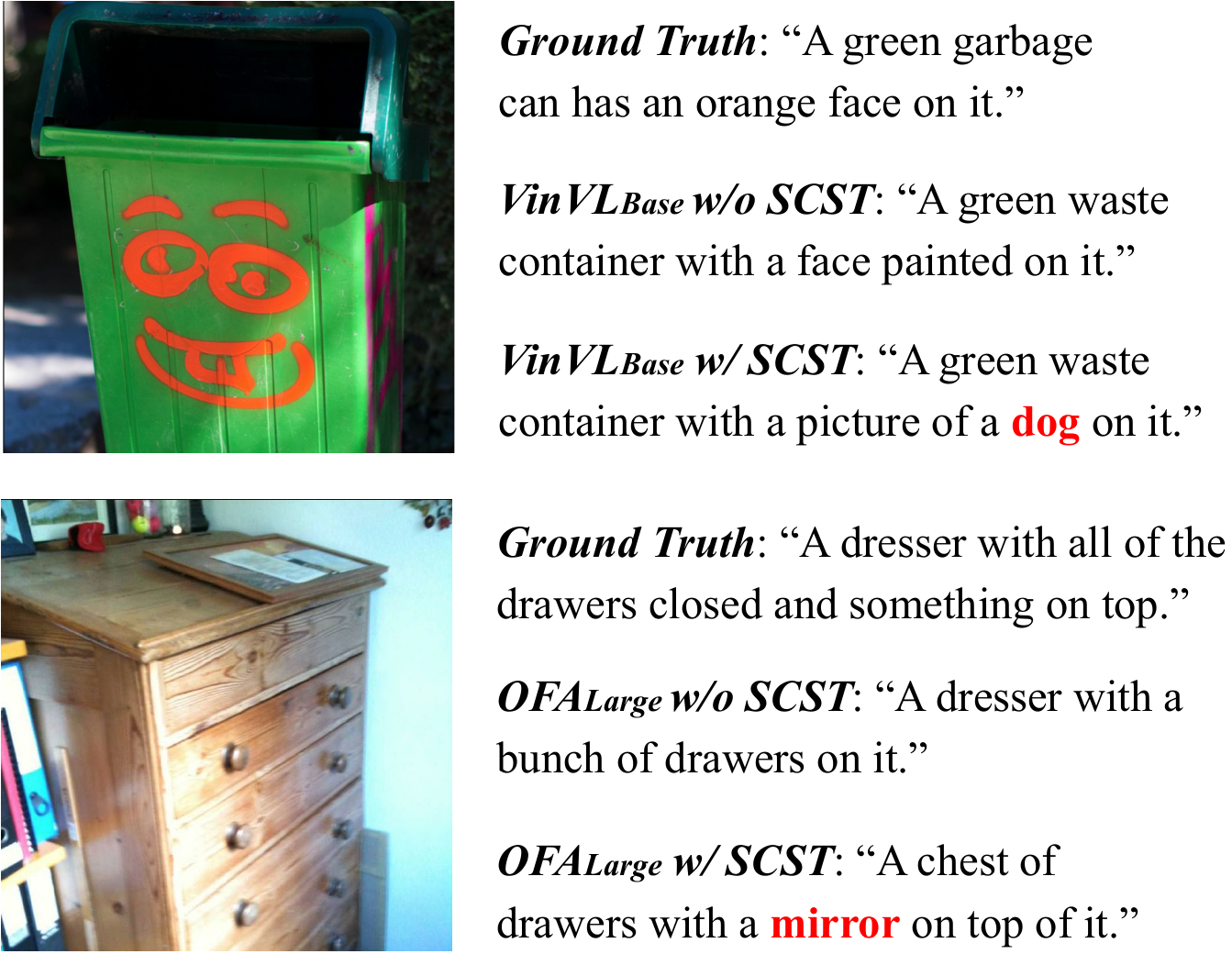}
    \caption{Comparison of image captioning examples generated by VinVL$_{Base}$ and OFA$_{Large}$ with and without the SCST CIDEr optimization. Red color denotes the occurrence of object hallucination.}
    \label{fig:scst_cast_study}
\end{figure}

Overall, we observe two noteworthy insights. 
Firstly, similar to the findings in \citet{rohrbach2018object}, for all CHAIR scores, they are not proportional to standard evaluation metrics. Although standard metrics (e.g., the cosine similarity in CIDEr) could potentially penalize the wrong object prediction, they do not directly reflect faithfulness. Captions can still have good scores from standard metrics as long as they contain sufficient accurate objects to fulfill coverage, even if hallucinated objects exist. For example, VinVL$_{Large}$ achieves higher CIDEr and BLEU-4 scores than VinVL$_{Base}$, but its CHAIR scores are also higher. Therefore, it is important to have a supplementary metric like CHAIR to reflect faithfulness besides other metrics.


Secondly, the Self-Critical Sequence Training (SCST)~\citep{Rennie2017SelfCriticalST} for the CIDEr optimization method harms the faithfulness of generated captions. SCST is a reinforcement learning algorithm that has been widely adopted as the second-stage finetuning after the standard cross-entropy optimization for image captioning~\citep{Anderson2018BottomUpAT,Zhou2020UnifiedVP,Li2020OscarOA,Zhang2021VinVLRV,Hu2021ScalingUV,Wang2022UnifyingAT}. It calculates the reward based on the CIDEr score by sampling captions during training without the need of another baseline.
Although SCST can significantly boost performance on previous standard metrics, it encourages models to generate more hallucinated objects in the captions.
For example, applying SCST improves the CIDEr score by 11.1 and BLEU-4 score by 2.7 for VinVL$_{Base}$, yet it also increases 0.9 CHAIR$_s$ score on the COCO Caption dataset. 

While \citet{Rennie2017SelfCriticalST} also observed this phenomenon by testing small scale models, we show that SCST hurts VLP models less. When the model is pre-trained very well, the side effect of SCST is alleviated (e.g., the OFA large model). Moreover, we demonstrate that this problem becomes more serious on out-of-domain images. For the VinVL$_{Base}$ model, there are 10.9\% more generated captions containing at least one hallucinated object after using SCST. We speculate that the CIDEr-based optimization encourages models to generate more words or phrases that have higher cosine similarities to the ground truth captions in the vision-language representation space, which can be plausible but not faithful. 

We show a case study in Figure~\ref{fig:scst_cast_study}. After finetuned by SCST, models will take a bigger risk to generate more detailed yet incorrect information (e.g., in the second example in Figure~\ref{fig:scst_cast_study}, the sentence with hallucination generates the detailed information ``mirror'', which cannot be found in the image). 
This will further amplify the object hallucination problem on out-of-domain images, as models may have lower confidence in unfamiliar visual concepts.


\section{Probing Image Encoding Methods and VLP Objectives} \label{sec:method} 

In this section, we systematically study two determinants in VLP that are intuitively influential to the severity of the object hallucination problem.
Firstly, we study how different types of image encoding affect object hallucination, as they are the key components of models to interpret visual information. Specifically, we ablate three encoding approaches including region-based, grid-based, and patch-based. Secondly, we analyze how different VLP objectives influence object hallucination. We ablate three commonly used ones: image-text contrastive (ITC), image-text matching (ITM), and image-conditioned language modeling (ICLM). 
Implementation details are described in Appendix~\ref{appendix:implementation_details}.


\subsection{Model Architecture} \label{sec:model_architecture}

\paragraph{CLIP.} CLIP~\citep{Radford2021LearningTV} is a dual-stream VLP model that consists of an image encoder and a text encoder. It is pre-trained on 400 million image-text pairs data using a cross-modal contrastive loss. Specifically, 
CLIP explores the image encoder with different sizes of two architectures\footnote{\url{https://github.com/openai/CLIP/blob/main/model-card.md}}, including the ResNet~\citep{He2016DeepRL} and the Vision Transformer (ViT)~\citep{Dosovitskiy2021AnII}. The resulting image and text encoders are aligned in the same multimodal feature space.

\paragraph{BERT.} BERT~\citep{Devlin2019BERTPO} is a Transformer~\citep{Vaswani2017AttentionIA} model pre-trained on a large corpus by the masked language modeling (MLM) and sentence permutation losses. It is shown to have excellent performance on various downstream tasks after finetuning. Moreover, BERT can also handle generation tasks when the self-attention layers are restricted to the left-to-right direction to generate text auto-regressively. In this paper, we refer to this variant as \text{BertLM}.

We design a flexible architecture that can plug in various visual encoders and fit modern VLP objectives without introducing extra influential factors.
As shown in Figure~\ref{fig:model_architecture}, the model consists of two parts, a visual encoder to encode images and a text decoder to generate sentences conditioned on the image representations. We use two separate modules rather than a unified single-stream model, as it is convenient to alter the visual encoder while keeping the text decoder the same. 
Specifically, for region-based image features, we explore the Faster R-CNN object detector~\citep{Ren2015FasterRT} with two different backbones: the ResNet-101 used in BUTD~\citep{Anderson2018BottomUpAT} and the ResNeXt-152~\citep{Xie2017AggregatedRT} used by \citet{Zhang2021VinVLRV}. They are both pre-trained on COCO~\citep{Lin2014MicrosoftCC} and Visual Genome~\citep{Krishna2016VisualGC} datasets for object detection.
For the grid-based and patch-based image features, we use the CLIP ResNet variants and CLIP ViT variants, respectively.
The reason for using CLIP is that all its variants are pre-trained on the same data and there is a wide range of different model sizes.
For all visual encoders, we use the same BertLM as the text decoder.

\begin{table}[]
\centering
\resizebox{\linewidth}{!}{%
\begin{tabular}{l|c|ccc|ccc}
\toprule
\multirow{2}{*}{\begin{tabular}[c]{@{}l@{}}Visual\\ Encoder\end{tabular}} & \multirow{2}{*}{\#Params} & \multicolumn{3}{c|}{\begin{tabular}[c]{@{}c@{}}COCO\\ Karpathy Test\end{tabular}} & \multicolumn{3}{c}{\begin{tabular}[c]{@{}c@{}}NoCaps Val\\ Out-of-domain\end{tabular}} \\
 &   & C$\uparrow$ & {CH$_i$}$\downarrow$ & {CH$_s$}$\downarrow$ & C$\uparrow$ & {CH$_i$}$\downarrow$ & {CH$_s$}$\downarrow$ \\ \midrule \midrule
\small{\textit{Region features}} &   &  &  &  &  &  &  \\
\multicolumn{1}{l|}{BUTD-RN101} & 45M & 110.6  & 9.1 & 15.9 & 40.5 & 36.7 & 49.0 \\
\multicolumn{1}{l|}{ResNeXt-152} & 60M & 115.9  & 7.1 & 12.9 & 45.1 & 30.5 & 41.1 \\ \midrule
\small{\textit{Grid features}} &  &   &  &  &  &  &  \\
\multicolumn{1}{l|}{RN50$\times$4} & 83M &  107.6 & 11.2 & 19.1 & 41.6 & 37.5 & 49.9  \\
\multicolumn{1}{l|}{RN50$\times$16} & 160M &  111.6 & 9.0 & 15.8 & 47.5 & 33.1 & 45.2 \\
\multicolumn{1}{l|}{RN50$\times$64} & 401M & 115.8 & 7.5 & 13.2 & 56.2 & 26.3 & 36.6 \\ \midrule
\small{\textit{Patch features}} &   &  &  &  &  &  &  \\
\multicolumn{1}{l|}{ViT-B/32} & 84M &  108.9 & 10.3 & 17.9 & 44.4 & 34.7 & 46.8 \\
\multicolumn{1}{l|}{ViT-B/16} & 82M &  111.8 & 8.1 & 14.7 & 51.9 & 30.3 & 42.3 \\
\multicolumn{1}{l|}{ViT-L/14} & 290M & 120.7 & 6.4 & 11.6 & 59.8 & 24.2 & 33.5 \\ \bottomrule
\end{tabular}%
}
\caption{Results of different types of visual encoders with the same BertLM text decoder on the COCO Karpathy test set and NoCaps validation set (out-of-domain).}
\label{tab:visual_encoder}
\end{table}

\subsection{Effects of Different Image Features} \label{sec:analysis_visual_features}
Recognizing visual objects correctly is crucial for avoiding object hallucination. In Table~\ref{tab:visual_encoder}, we compare the 
performance of different visual encoders with the same text decoder on COCO (in-domain) and NoCaps (out-of-domain) datasets.


Overall, patch-based visual encoders attain the best performance in terms of avoiding object hallucination.
Models with grid features hallucinate more frequently when achieving comparable CIDEr scores to the other models. For example, on COCO, RN50$\times$16 has a similar CIDEr score to \text{ViT-B/16} but higher CHAIR$_s$, which is also observed between RN50$\times$64 and ResNeXt-152. We conjecture that the inductive biases~\citep{Cohen2017InductiveBO} of the Convolutional Neural Network (CNN), such as locality and translation invariance, weaken the connection of different characteristics of a single object and thus lead to more hallucination. Oppositely, regional or patch-level features are obtained by directly dividing images into different parts and further associating them through positional embeddings. In addition, we see that a smaller patch resolution helps to reduce object hallucination without enlarging the model size. 

For region-based visual encoders, although they achieve modest results on COCO with relatively small model sizes, their performance of object hallucination on out-of-domain images drops dramatically. One important reason is that the output of such encoders only contains representations of detected visual objects rather than the whole image, which may amplify detection errors as there is much less context. Moreover, as the object detector is pre-trained separately from the whole model and its parameters are fixed during finetuning, this gap could also aggravate object hallucination on unseen images.



\subsection{Effects of Different VLP Objectives} \label{sec:analysis_vlp_objectives}
Based on the best performing ViT-L/14 baseline, we explore three commonly used vision-language pre-training objectives and their variants that could potentially affect object hallucination. 

\subsubsection{Pre-training Datasets}
We explore two pre-training datasets with image-text pairs: 1) the VG Caption from the Visual Genome~\citep{Krishna2016VisualGC} dataset, which contains 10K images and each image has multiple corresponding descriptions; and 2) a more large-scale dataset CC3M~\citep{Sharma2018ConceptualCA} that contains three millions of image-text pairs.

\begin{table}[]
\centering
\resizebox{\linewidth}{!}{%
\setlength{\tabcolsep}{2.55mm}{
\begin{tabular}{l|ccc|ccc}
\toprule
\multirow{2}{*}{\begin{tabular}[c]{@{}l@{}}VLP Objectives\end{tabular}} & \multicolumn{3}{c|}{\begin{tabular}[c]{@{}c@{}}COCO\\ Karpathy Test\end{tabular}} & \multicolumn{3}{c}{\begin{tabular}[c]{@{}c@{}}NoCaps Val\\ Out-of-domain\end{tabular}} \\
 & C$\uparrow$ & {CH$_i$}$\downarrow$ & {CH$_s$}$\downarrow$ & C$\uparrow$ & {CH$_i$}$\downarrow$ & {CH$_s$}$\downarrow$ \\ \midrule \midrule
(a) None & 120.7 & 6.4 & 11.6 & 59.8 & 24.2 & 33.5 \\ \midrule
\multicolumn{7}{c}{\textit{Discriminative Objectives}} \\ \midrule
\multicolumn{1}{l|}{\small{\textit{CC3M}}} &   &  &  &  &  &  \\
\multicolumn{1}{l|}{(b) ITC} & 120.5 & 6.5 & 11.7 & 59.9 & 24.4 & 33.8 \\ 
\multicolumn{1}{l|}{(c) ITC$_\textit{Late}$} & 121.2 & 6.2 & 11.3 & 60.5 & 23.8 & 32.9 \\
\multicolumn{1}{l|}{(d) ITC$_\textit{Late}$ + ITM} & 121.0 & 6.3 & 11.5 & 60.2 & 23.9 & 33.1 \\ 
\multicolumn{1}{l|}{(e) ITC$_\textit{Late}$ + ITM$_\textit{Hard}$} & 120.9 & 6.6 & 11.7 & 59.9 & 24.2 & 33.3 \\ \midrule
\multicolumn{7}{c}{\textit{Generative Objectives}} \\ \midrule
\multicolumn{1}{l|}{\small{\textit{Visual Genome}}} &   &  &  &  &  &  \\
\multicolumn{1}{l|}{(f) LM} & 120.3 & 5.5 & 9.8 & 62.8 & 9.0 & 13.9 \\ 
\multicolumn{1}{l|}{(g) LM + ObjectMLM} & 121.9 & 5.3 & 9.2 & 63.8 & 8.8 & 13.1 \\ \midrule
\multicolumn{1}{l|}{\small{\textit{CC3M}}} &   &  &  &  &  &  \\
\multicolumn{1}{l|}{(h) LM} & 122.3 & 6.0 & 10.9 & 92.1 & 8.3 & 14.5 \\ 
\multicolumn{1}{l|}{(i) LM + ObjectMLM} & 124.5 & 5.1 & 9.0 & 94.0 & 8.0 & 13.1 \\ \midrule
\multicolumn{1}{l|}{(c) + (i)} & \textbf{125.1} & \textbf{4.9} & \textbf{8.8} & \textbf{94.5} & \textbf{7.9} & \textbf{12.5} \\ \bottomrule
\end{tabular}%
}
}
\caption{Comparison of the effects of different VLP objectives and their combination on object hallucination.}
\label{tab:vlp_objectives}
\end{table}

\subsubsection{Image-Text Contrastive (ITC) Loss}
The cross-modal contrastive loss is shown to be fairly effective in representation learning~\citep{Tian2020ContrastiveMC,Sigurdsson2020VisualGI} and VLP~\citep{Radford2021LearningTV,Li2021AlignBF,Li2022BLIPBL}. It aligns the visual and textual representations into the same multimodal feature space by shortening the distance between an image and a text if they are paired, or enlarging if they are not. 

Counter-intuitively, as shown in Table~\ref{tab:vlp_objectives} (b), ITC has negligible influence on the faithfulness of generated captions. We speculate that it only enhances the model's understanding of global-level representations rather than token-level alignment between images and texts. To verify, we further test the ITC with a more fine-grained token-level late interaction (ITC$_\textit{Late}$) proposed by \citet{Yao2021FILIPFI}. As shown in Table~\ref{tab:vlp_objectives} (c), ITC$_\textit{Late}$ is more effective than the vanilla ITC and slightly reduces object hallucination. We think this benefits from the word-patch alignment ability enabled by ITC$_\textit{Late}$.

\subsubsection{Image-Text Matching (ITM) Loss}
ITM is a widely used objective in VLP~\citep{Li2020UnicoderVLAU,Chen2020UNITERUI,Zhou2021UC2UC}. It is a binary classification task that aims to make the model learn whether an image and a sentence are paired or not. Based on that, ITM with hard negatives (ITM$_\textit{Hard}$) is introduced to increase the difficulty of the task, which is shown to be very effective on representation learning~\citep{Kalantidis2020HardNM,Robinson2021ContrastiveLW,Li2021UNIMOTU}. We follow the ITM loss proposed by \citet{Li2022BLIPBL}, in which an in-batch negative example is sampled either uniformly (normal negative) or from the similarity distribution of image-text pairs computed by ITC (hard negative).

The results are exhibited in Table~\ref{tab:vlp_objectives} (d) (e). 
Both ITM and ITM$_\textit{Hard}$ are not highly correlated with the object hallucination problem. They only slightly reduce hallucination in generated texts on out-of-domain images. Although the ITM$_\textit{Hard}$ can be seen as an analogy to the object hallucination problem (plausible but not correct) in a global and discriminative way, it has a negligible effect on reducing hallucination for downstream generative tasks.

\begin{figure}[t!]
    \centering
    \includegraphics[width=\linewidth]{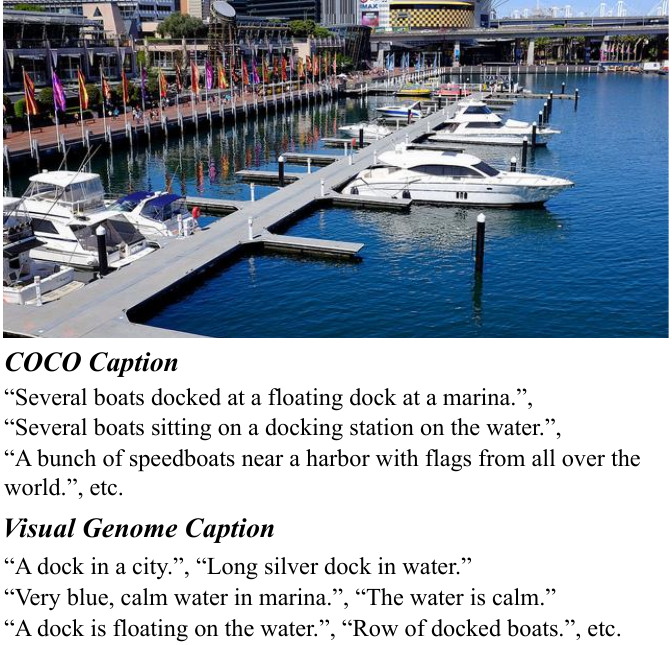}
    \caption{Comparison of ground truth captions in COCO and Visual Genome datasets for the same image.}
    \label{fig:vg_coco_compare}
\end{figure}

\begin{figure}[t!]
    \centering
    \includegraphics[width=\linewidth]{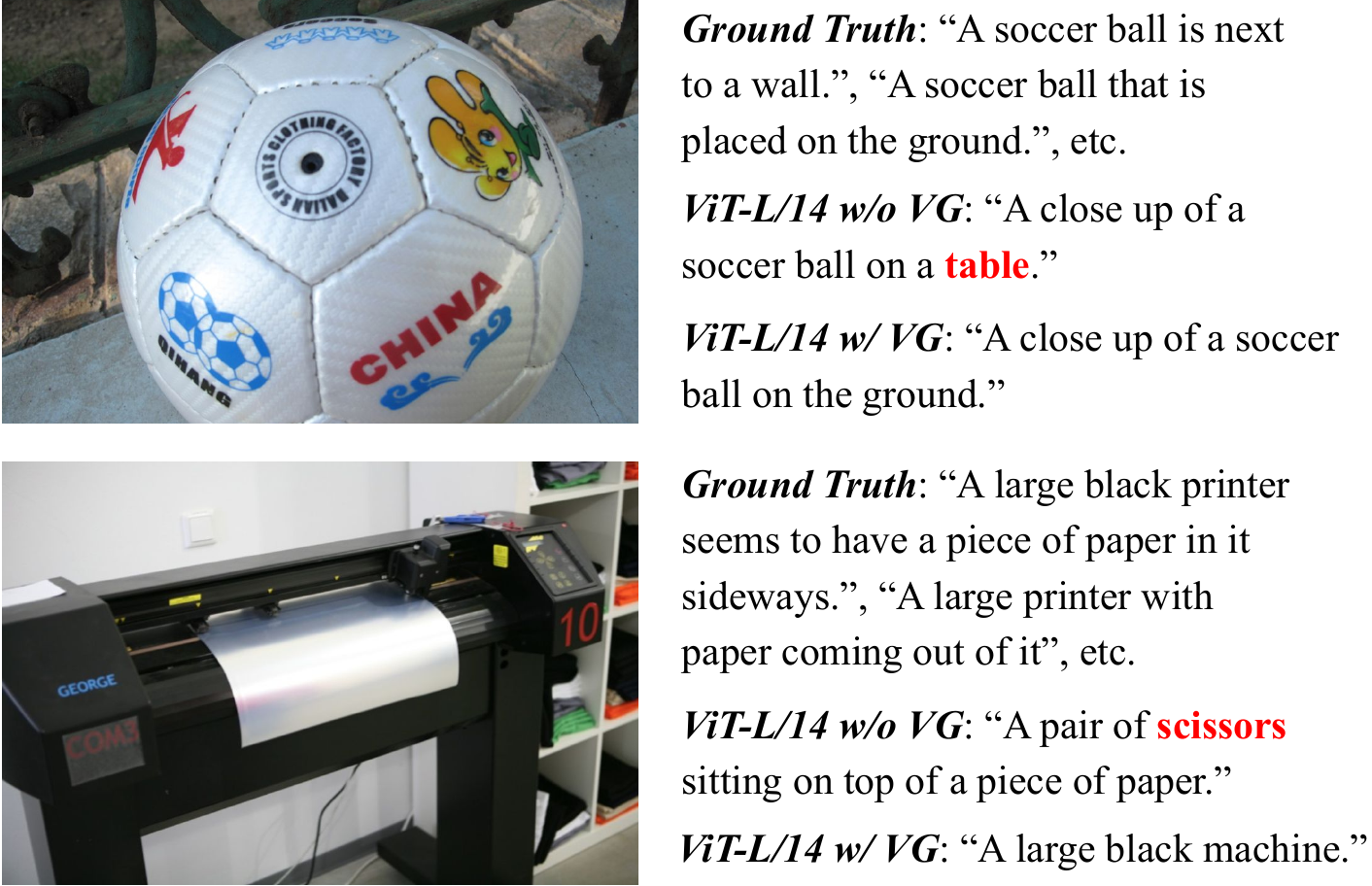}
    \caption{Comparison of generated captions with or without the image-conditioned language modeling pre-training on the VG dataset before finetuning.}
    \label{fig:case_vg}
\end{figure}

\subsubsection{Image-Conditioned Language Modeling}
Various image-conditioned language modeling losses have been proposed in the VLP research, in the form of masked language modeling (MLM)~\citep{Sun2019VideoBERTAJ,Tan2019LXMERTLC,Su2020VLBERTPO}, text infilling~\citep{dai-etal-2022-enabling,Wang2022UnifyingAT}, prefix LM~\citep{Wang2021SimVLMSV}, and causal LM~\citep{Hu2021ScalingUV}. This is one of the most crucial pre-training losses to activate the cross-modal text generation ability for the VLP model. 

We first examine the causal LM loss, which is exactly the same loss as the image captioning loss, but used in the pre-training on a much larger scale. Surprisingly, as shown in Table~\ref{tab:vlp_objectives} (f), although pre-training on the VG Caption does not improve previous standard metrics like CIDEr, it helps to reduce object hallucination by a large margin when compared to (a). 

There are two reasons behind this performance lift. Firstly, as described in Figure~\ref{fig:vg_coco_compare}, for each image, VG contains more captions than COCO. Each caption in VG is much shorter and only describes one specific aspect of the image, unlike the global descriptions in COCO. Therefore, pre-training on VG and then finetuning on COCO is a fine-to-coarse process. It enables models to first accurately describe different parts of an image and connect these clues together at a higher viewing point. 
Secondly, due to the nature of the short length of VG captions, the model becomes slightly more cautious. On average, after adding VG data in the pre-training, there are 0.08 and 0.24 fewer objects generated in each caption on COCO and NoCaps, respectively.
This observation aligns with the sentence simplification method proposed by \citet{biten2022let}, which simplifies sentences to augment data and further mitigate object hallucination.
Figure~\ref{fig:case_vg} illustrates VG's effects on generated samples. The model is more faithful but more likely to lack some details when it is not confident. 

For CC3M, we observe a leap in all metrics. It improves the general image translation ability of the model, which can be seen as large-scale data augmentation. This indicates that seeing a sufficient amount of data and co-occurrence of various objects during pre-training help to mitigate object hallucination to some extent. However, data augmentation may not be the key to drastically tackle object hallucination. As discussed in Section~\ref{sec:sota_models}, object hallucination still happens frequently even if the model is pre-trained on large-scale data. Therefore, we believe that enhancing the controllability of vision-conditioned text generation would be a promising future direction. More case studies are exhibited in Appendix~\ref{appendix:additional_cases}.

\section{Object Masked Language Modeling}

\begin{figure}[t!]
    \centering
    \includegraphics[width=\linewidth]{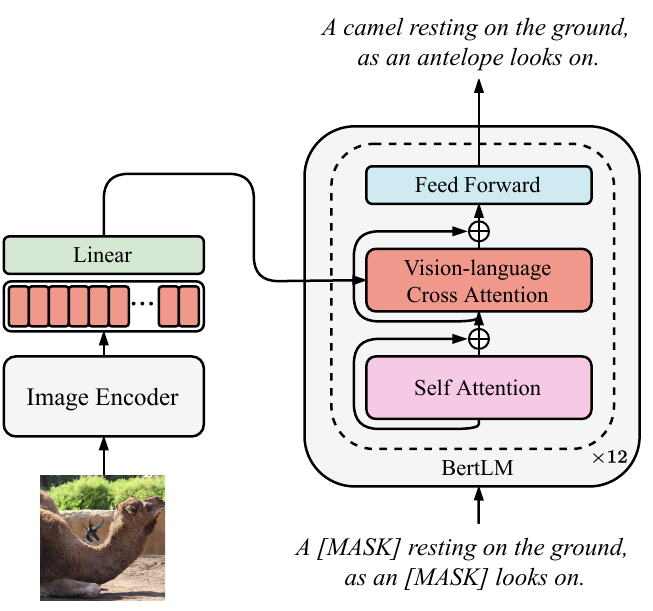}
    \caption{An overview of the model architecture and the training of our proposed ObjMLM. We use the same architecture as described in Section~\ref{sec:method} to show the effectiveness of ObjMLM. Here, the image encoder can be one of the region-based, grid-based, or patch-based variants as described in Section~\ref{sec:analysis_visual_features}. For ObjMLM, we use the ViT-L/14.}
    \label{fig:model_architecture}
\end{figure}

Based on the findings in Section~\ref{sec:method}, we propose a simple yet effective pre-training objective to mitigate object hallucination by improving object-level image-text alignment. It is named Object Masked Language Modeling (ObjMLM).
As shown in Figure~\ref{fig:model_architecture}, ObjMLM can be seen as a variant of the MLM loss by masking all the objects in the text that appear in the image. 
For each sentence, we mask the object words and phrases as defined in the object category lists of both COCO and NoCaps by performing exact matching. Similar to the whole word masking~\citep{Cui2021PreTrainingWW}, we conduct whole object masking so that there will be only one \texttt{[MASK]} token to replace each object.

Compare the results shown in lines (h) and (i) of Table~\ref{tab:vlp_objectives}, by plugging ObjMLM into an existing VLP setting, the CHAIR$_s$ score is reduced by 17.4\%. This is a non-trivial improvement without introducing more pre-training data. To further validate ObjMLM's effectiveness, we replace it by the standard MLM loss with a 15\% masking rate. However, it only reduces CHAIR$_s$ by 1.7\%, which is not significant.
We conjecture that ObjMLM adds a constraint that indirectly controls the model to only generate objects that are visible in the input image.
Additionally, ObjMLM enhances the model's recognition ability when describing the spatial relationship between objects, which is a common scenario that causes hallucinations frequently. 

\section{Conclusion} \label{sec:conclusion}

This paper systematically studies the objection hallucination phenomenon in VLP models, which is a severe problem but neglected in contemporary VLP works. We find that recent large VLP models still hallucinate frequently. Moreover, the widely used SCST method harms the faithfulness of generated sentences in image captioning, even if it improves previous standard metrics. Furthermore, we discover that image encoding matters and the patch-based input with smaller resolution helps mitigate object hallucination. Finally, we ablate commonly used VLP losses and show that token-level image-text alignment and controllability of the generation are crucial. We further propose a new loss named ObjMLM, which reduces object hallucination by 17.4\% for an existing VLP setting.
We believe our findings are beneficial for future work to build more responsible VLP models. 

\section*{Limitations}
We understand that the hallucination problem is a big research topic and it is not just limited to object hallucination. In this paper, we focus on the investigation and mitigation of object hallucination, leaving other types of hallucination in VLP for future work. Another limitation is that for the discussion of recent VLP models in Section~\ref{sec:sota_models}, we only study those whose pre-trained checkpoints are publicly available. For the non-released ones, we cannot pre-train them by ourselves due to the lack of large-scale GPU power and private pre-training datasets.

\bibliography{custom}

\begin{thebibliography}{70}
\expandafter\ifx\csname natexlab\endcsname\relax\def\natexlab#1{#1}\fi

\bibitem[{Agrawal et~al.(2019)Agrawal, Desai, Wang, Chen, Jain, Johnson, Batra,
  Parikh, Lee, and Anderson}]{Agrawal2019nocapsNO}
Harsh Agrawal, Karan Desai, Yufei Wang, Xinlei Chen, Rishabh Jain, Mark
  Johnson, Dhruv Batra, Devi Parikh, Stefan Lee, and Peter Anderson. 2019.
\newblock nocaps: novel object captioning at scale.
\newblock \emph{International Conference on Computer Vision}, pages 8947--8956.

\bibitem[{Alayrac et~al.(2022)Alayrac, Donahue, Luc, Miech, Barr, Hasson, Lenc,
  Mensch, Millican, Reynolds, Ring, Rutherford, Cabi, Han, Gong, Samangooei,
  Monteiro, Menick, Borgeaud, Brock, Nematzadeh, Sharifzadeh, Binkowski,
  Barreira, Vinyals, Zisserman, and Simonyan}]{Alayrac2022FlamingoAV}
Jean-Baptiste Alayrac, Jeff Donahue, Pauline Luc, Antoine Miech, Iain Barr,
  Yana Hasson, Karel Lenc, Arthur Mensch, Katherine Millican, Malcolm Reynolds,
  Roman Ring, Eliza Rutherford, Serkan Cabi, Tengda Han, Zhitao Gong, Sina
  Samangooei, Marianne Monteiro, Jacob Menick, Sebastian Borgeaud, Andrew
  Brock, Aida Nematzadeh, Sahand Sharifzadeh, Mikolaj Binkowski, Ricardo
  Barreira, Oriol Vinyals, Andrew Zisserman, and Karen Simonyan. 2022.
\newblock \href {https://openreview.net/forum?id=EbMuimAbPbs} {Flamingo: a
  visual language model for few-shot learning}.
\newblock In \emph{Advances in Neural Information Processing Systems}.

\bibitem[{Anderson et~al.(2016)Anderson, Fernando, Johnson, and
  Gould}]{Anderson2016SPICESP}
Peter Anderson, Basura Fernando, Mark Johnson, and Stephen Gould. 2016.
\newblock Spice: Semantic propositional image caption evaluation.
\newblock In \emph{ECCV}.

\bibitem[{Anderson et~al.(2018)Anderson, He, Buehler, Teney, Johnson, Gould,
  and Zhang}]{Anderson2018BottomUpAT}
Peter Anderson, Xiaodong He, Chris Buehler, Damien Teney, Mark Johnson, Stephen
  Gould, and Lei Zhang. 2018.
\newblock Bottom-up and top-down attention for image captioning and visual
  question answering.
\newblock \emph{2018 IEEE/CVF Conference on Computer Vision and Pattern
  Recognition}, pages 6077--6086.

\bibitem[{Banerjee and Lavie(2005)}]{Banerjee2005METEORAA}
Satanjeev Banerjee and Alon Lavie. 2005.
\newblock Meteor: An automatic metric for mt evaluation with improved
  correlation with human judgments.
\newblock In \emph{IEEvaluation@ACL}.

\bibitem[{Biten et~al.(2022)Biten, G\'omez, and Karatzas}]{biten2022let}
Ali~Furkan Biten, Llu{\'\i}s G\'omez, and Dimosthenis Karatzas. 2022.
\newblock Let there be a clock on the beach: Reducing object hallucination in
  image captioning.
\newblock In \emph{Proceedings of the IEEE/CVF Winter Conference on
  Applications of Computer Vision (WACV)}, pages 1381--1390.

\bibitem[{Chen et~al.(2020)Chen, Li, Yu, Kholy, Ahmed, Gan, Cheng, and
  Liu}]{Chen2020UNITERUI}
Yen-Chun Chen, Linjie Li, Licheng Yu, Ahmed~El Kholy, Faisal Ahmed, Zhe Gan,
  Yu~Cheng, and Jingjing Liu. 2020.
\newblock Uniter: Universal image-text representation learning.
\newblock In \emph{ECCV}.

\bibitem[{Cho et~al.(2021)Cho, Lei, Tan, and Bansal}]{Cho2021UnifyingVT}
Jaemin Cho, Jie Lei, Hao Tan, and Mohit Bansal. 2021.
\newblock \href {https://proceedings.mlr.press/v139/cho21a.html} {Unifying
  vision-and-language tasks via text generation}.
\newblock In \emph{Proceedings of the 38th International Conference on Machine
  Learning}, volume 139 of \emph{Proceedings of Machine Learning Research},
  pages 1931--1942. PMLR.

\bibitem[{Cohen and Shashua(2017)}]{Cohen2017InductiveBO}
Nadav Cohen and Amnon Shashua. 2017.
\newblock \href {https://openreview.net/forum?id=BkVsEMYel} {Inductive bias of
  deep convolutional networks through pooling geometry}.
\newblock In \emph{International Conference on Learning Representations}.

\bibitem[{Cui et~al.(2021)Cui, Che, Liu, Qin, Yang, Wang, and
  Hu}]{Cui2021PreTrainingWW}
Yiming Cui, Wanxiang Che, Ting Liu, Bing Qin, Ziqing Yang, Shijin Wang, and
  Guoping Hu. 2021.
\newblock Pre-training with whole word masking for chinese bert.
\newblock \emph{IEEE/ACM Transactions on Audio, Speech, and Language
  Processing}, 29:3504--3514.

\bibitem[{Dai et~al.(2022)Dai, Hou, Shang, Jiang, Liu, and
  Fung}]{dai-etal-2022-enabling}
Wenliang Dai, Lu~Hou, Lifeng Shang, Xin Jiang, Qun Liu, and Pascale Fung. 2022.
\newblock \href {https://doi.org/10.18653/v1/2022.findings-acl.187} {Enabling
  multimodal generation on {CLIP} via vision-language knowledge distillation}.
\newblock In \emph{Findings of the Association for Computational Linguistics:
  ACL 2022}, pages 2383--2395, Dublin, Ireland. Association for Computational
  Linguistics.

\bibitem[{Devlin et~al.(2019)Devlin, Chang, Lee, and
  Toutanova}]{Devlin2019BERTPO}
Jacob Devlin, Ming-Wei Chang, Kenton Lee, and Kristina Toutanova. 2019.
\newblock \href {https://doi.org/10.18653/v1/N19-1423} {{BERT}: Pre-training of
  deep bidirectional transformers for language understanding}.
\newblock In \emph{Proceedings of the 2019 Conference of the North {A}merican
  Chapter of the Association for Computational Linguistics: Human Language
  Technologies, Volume 1 (Long and Short Papers)}, pages 4171--4186,
  Minneapolis, Minnesota. Association for Computational Linguistics.

\bibitem[{Ding et~al.(2021)Ding, Yang, Hong, Zheng, Zhou, Yin, Lin, Zou, Shao,
  Yang, and Tang}]{Ding2021CogViewMT}
Ming Ding, Zhuoyi Yang, Wenyi Hong, Wendi Zheng, Chang Zhou, Da~Yin, Junyang
  Lin, Xu~Zou, Zhou Shao, Hongxia Yang, and Jie Tang. 2021.
\newblock Cogview: Mastering text-to-image generation via transformers.
\newblock In \emph{NeurIPS}.

\bibitem[{Dosovitskiy et~al.(2021)Dosovitskiy, Beyer, Kolesnikov, Weissenborn,
  Zhai, Unterthiner, Dehghani, Minderer, Heigold, Gelly, Uszkoreit, and
  Houlsby}]{Dosovitskiy2021AnII}
Alexey Dosovitskiy, Lucas Beyer, Alexander Kolesnikov, Dirk Weissenborn,
  Xiaohua Zhai, Thomas Unterthiner, Mostafa Dehghani, Matthias Minderer, Georg
  Heigold, Sylvain Gelly, Jakob Uszkoreit, and Neil Houlsby. 2021.
\newblock \href {https://openreview.net/forum?id=YicbFdNTTy} {An image is worth
  16x16 words: Transformers for image recognition at scale}.
\newblock In \emph{International Conference on Learning Representations}.

\bibitem[{Dziri et~al.(2021)Dziri, Madotto, Zaiane, and Bose}]{dziri2021neural}
Nouha Dziri, Andrea Madotto, Osmar Zaiane, and Avishek~Joey Bose. 2021.
\newblock Neural path hunter: Reducing hallucination in dialogue systems via
  path grounding.
\newblock \emph{EMNLP}.

\bibitem[{He et~al.(2016)He, Zhang, Ren, and Sun}]{He2016DeepRL}
Kaiming He, X.~Zhang, Shaoqing Ren, and Jian Sun. 2016.
\newblock Deep residual learning for image recognition.
\newblock \emph{2016 IEEE Conference on Computer Vision and Pattern Recognition
  (CVPR)}, pages 770--778.

\bibitem[{Hu et~al.(2022)Hu, Gan, Wang, Yang, Liu, Lu, and
  Wang}]{Hu2021ScalingUV}
X.~Hu, Z.~Gan, J.~Wang, Z.~Yang, Z.~Liu, Y.~Lu, and L.~Wang. 2022.
\newblock \href {https://doi.org/10.1109/CVPR52688.2022.01745} {Scaling up
  vision-language pretraining for image captioning}.
\newblock In \emph{2022 IEEE/CVF Conference on Computer Vision and Pattern
  Recognition (CVPR)}, pages 17959--17968, Los Alamitos, CA, USA. IEEE Computer
  Society.

\bibitem[{Ji et~al.(2022)Ji, Lee, Frieske, Yu, Su, Xu, Ishii, Bang, Dai,
  Madotto, and Fung}]{ji2022survey}
Ziwei Ji, Nayeon Lee, Rita Frieske, Tiezheng Yu, Dan Su, Yan Xu, Etsuko Ishii,
  Yejin Bang, Wenliang Dai, Andrea Madotto, and Pascale Fung. 2022.
\newblock Survey of hallucination in natural language generation.
\newblock \emph{ACM Computing Surveys}.

\bibitem[{Jia et~al.(2021)Jia, Yang, Xia, Chen, Parekh, Pham, Le, Sung, Li, and
  Duerig}]{Jia2021ScalingUV}
Chao Jia, Yinfei Yang, Ye~Xia, Yi-Ting Chen, Zarana Parekh, Hieu Pham, Quoc~V.
  Le, Yun-Hsuan Sung, Zhen Li, and Tom Duerig. 2021.
\newblock Scaling up visual and vision-language representation learning with
  noisy text supervision.
\newblock In \emph{ICML}.

\bibitem[{Kalantidis et~al.(2020)Kalantidis, Sariyildiz, Pion, Weinzaepfel, and
  Larlus}]{Kalantidis2020HardNM}
Yannis Kalantidis, Mert~Bulent Sariyildiz, Noe Pion, Philippe Weinzaepfel, and
  Diane Larlus. 2020.
\newblock \href
  {https://proceedings.neurips.cc/paper/2020/file/f7cade80b7cc92b991cf4d2806d6bd78-Paper.pdf}
  {Hard negative mixing for contrastive learning}.
\newblock In \emph{Advances in Neural Information Processing Systems},
  volume~33, pages 21798--21809. Curran Associates, Inc.

\bibitem[{Karpathy and Fei-Fei(2017)}]{Karpathy2017DeepVA}
Andrej Karpathy and Li~Fei-Fei. 2017.
\newblock Deep visual-semantic alignments for generating image descriptions.
\newblock \emph{IEEE Transactions on Pattern Analysis and Machine
  Intelligence}, 39:664--676.

\bibitem[{Kim et~al.(2021)Kim, Son, and Kim}]{Kim2021ViLTVT}
Wonjae Kim, Bokyung Son, and Ildoo Kim. 2021.
\newblock Vilt: Vision-and-language transformer without convolution or region
  supervision.
\newblock In \emph{ICML}.

\bibitem[{Krishna et~al.(2016)Krishna, Zhu, Groth, Johnson, Hata, Kravitz,
  Chen, Kalantidis, Li, Shamma, Bernstein, and Fei-Fei}]{Krishna2016VisualGC}
Ranjay Krishna, Yuke Zhu, Oliver Groth, Justin Johnson, Kenji Hata, Joshua
  Kravitz, Stephanie Chen, Yannis Kalantidis, Li-Jia Li, David~A. Shamma,
  Michael~S. Bernstein, and Li~Fei-Fei. 2016.
\newblock Visual genome: Connecting language and vision using crowdsourced
  dense image annotations.
\newblock \emph{International Journal of Computer Vision}, 123:32--73.

\bibitem[{Kuznetsova et~al.(2020)Kuznetsova, Rom, Alldrin, Uijlings, Krasin,
  Pont-Tuset, Kamali, Popov, Malloci, Kolesnikov, Duerig, and
  Ferrari}]{Kuznetsova2020TheOI}
Alina Kuznetsova, Hassan Rom, Neil~Gordon Alldrin, Jasper R.~R. Uijlings, Ivan
  Krasin, Jordi Pont-Tuset, Shahab Kamali, Stefan Popov, Matteo Malloci,
  Alexander Kolesnikov, Tom Duerig, and Vittorio Ferrari. 2020.
\newblock The open images dataset v4.
\newblock \emph{International Journal of Computer Vision}, 128:1956--1981.

\bibitem[{Li et~al.(2020{\natexlab{a}})Li, Duan, Fang, Jiang, and
  Zhou}]{Li2020UnicoderVLAU}
Gen Li, Nan Duan, Yuejian Fang, Daxin Jiang, and Ming Zhou. 2020{\natexlab{a}}.
\newblock Unicoder-vl: A universal encoder for vision and language by
  cross-modal pre-training.
\newblock In \emph{AAAI}.

\bibitem[{Li et~al.(2022)Li, Li, Xiong, and Hoi}]{Li2022BLIPBL}
Junnan Li, Dongxu Li, Caiming Xiong, and Steven Hoi. 2022.
\newblock Blip: Bootstrapping language-image pre-training for unified
  vision-language understanding and generation.
\newblock In \emph{ICML}.

\bibitem[{Li et~al.(2021{\natexlab{a}})Li, Selvaraju, Gotmare, Joty, Xiong, and
  Hoi}]{Li2021AlignBF}
Junnan Li, Ramprasaath~R. Selvaraju, Akhilesh~Deepak Gotmare, Shafiq~R. Joty,
  Caiming Xiong, and Steven C.~H. Hoi. 2021{\natexlab{a}}.
\newblock Align before fuse: Vision and language representation learning with
  momentum distillation.
\newblock In \emph{NeurIPS}.

\bibitem[{Li et~al.(2020{\natexlab{b}})Li, Yatskar, Yin, Hsieh, and
  Chang}]{Li2019VisualBERTAS}
Liunian~Harold Li, Mark Yatskar, Da~Yin, Cho-Jui Hsieh, and Kai-Wei Chang.
  2020{\natexlab{b}}.
\newblock \href {https://doi.org/10.18653/v1/2020.acl-main.469} {What does
  {BERT} with vision look at?}
\newblock In \emph{Proceedings of the 58th Annual Meeting of the Association
  for Computational Linguistics}, pages 5265--5275, Online. Association for
  Computational Linguistics.

\bibitem[{Li et~al.(2021{\natexlab{b}})Li, Gao, Niu, Xiao, Liu, Liu, Wu, and
  Wang}]{Li2021UNIMOTU}
Wei Li, Can Gao, Guocheng Niu, Xinyan Xiao, Hao Liu, Jiachen Liu, Hua Wu, and
  Haifeng Wang. 2021{\natexlab{b}}.
\newblock \href {https://doi.org/10.18653/v1/2021.acl-long.202} {{UNIMO}:
  Towards unified-modal understanding and generation via cross-modal
  contrastive learning}.
\newblock In \emph{Proceedings of the 59th Annual Meeting of the Association
  for Computational Linguistics and the 11th International Joint Conference on
  Natural Language Processing (Volume 1: Long Papers)}, pages 2592--2607,
  Online. Association for Computational Linguistics.

\bibitem[{Li et~al.(2020{\natexlab{c}})Li, Yin, Li, Hu, Zhang, Zhang, Wang, Hu,
  Dong, Wei, Choi, and Gao}]{Li2020OscarOA}
Xiujun Li, Xi~Yin, Chunyuan Li, Xiaowei Hu, Pengchuan Zhang, Lei Zhang, Lijuan
  Wang, Houdong Hu, Li~Dong, Furu Wei, Yejin Choi, and Jianfeng Gao.
  2020{\natexlab{c}}.
\newblock Oscar: Object-semantics aligned pre-training for vision-language
  tasks.
\newblock In \emph{ECCV}.

\bibitem[{Lin et~al.(2021)Lin, Men, Yang, Zhou, Ding, Zhang, Wang, Wang, Jiang,
  Jia, Zhang, Zhang, Zou, Li, Deng, Liu, Xue, Zhou, Ma, Yu, Li, Lin, Zhou,
  ie~Tang, and Yang}]{Lin2021M6AC}
Junyang Lin, Rui Men, An~Yang, Chan Zhou, Ming Ding, Yichang Zhang, Peng Wang,
  Ang Wang, Le~Jiang, Xianyan Jia, J.~Zhang, Jianwei Zhang, Xu~Zou, Zhikang Li,
  Xiao~Qing Deng, Jie Liu, Jinbao Xue, Huiling Zhou, Jianxin Ma, Jin Yu, Yong
  Li, Wei Lin, Jingren Zhou, J~ie~Tang, and Hongxia Yang. 2021.
\newblock M6: A chinese multimodal pretrainer.
\newblock \emph{ArXiv}, abs/2103.00823.

\bibitem[{Lin et~al.(2014)Lin, Maire, Belongie, Hays, Perona, Ramanan,
  Doll{\'a}r, and Zitnick}]{Lin2014MicrosoftCC}
Tsung-Yi Lin, Michael Maire, Serge~J. Belongie, James Hays, Pietro Perona, Deva
  Ramanan, Piotr Doll{\'a}r, and C.~Lawrence Zitnick. 2014.
\newblock Microsoft coco: Common objects in context.
\newblock In \emph{ECCV}.

\bibitem[{Loshchilov and Hutter(2019)}]{Loshchilov2019DecoupledWD}
Ilya Loshchilov and Frank Hutter. 2019.
\newblock Decoupled weight decay regularization.
\newblock In \emph{ICLR}.

\bibitem[{Lu et~al.(2019)Lu, Batra, Parikh, and Lee}]{Lu2019ViLBERTPT}
Jiasen Lu, Dhruv Batra, Devi Parikh, and Stefan Lee. 2019.
\newblock Vilbert: Pretraining task-agnostic visiolinguistic representations
  for vision-and-language tasks.
\newblock In \emph{NeurIPS}.

\bibitem[{Lu et~al.(2018)Lu, Yang, Batra, and Parikh}]{Lu2018NeuralBT}
Jiasen Lu, Jianwei Yang, Dhruv Batra, and Devi Parikh. 2018.
\newblock Neural baby talk.
\newblock \emph{2018 IEEE/CVF Conference on Computer Vision and Pattern
  Recognition}, pages 7219--7228.

\bibitem[{Ma et~al.(2020)Ma, Kalantidis, AlRegib, Vajda, Rohrbach, and
  Kira}]{ma2020learning}
Chih-Yao Ma, Yannis Kalantidis, Ghassan AlRegib, Peter Vajda, Marcus Rohrbach,
  and Zsolt Kira. 2020.
\newblock \href {https://arxiv.org/abs/1906.00283} {Learning to generate
  grounded image captions without localization supervision}.
\newblock In \emph{Proceedings of the European Conference on Computer Vision
  (ECCV)}.

\bibitem[{Maynez et~al.(2020)Maynez, Narayan, Bohnet, and
  McDonald}]{maynez2020faithfulness}
Joshua Maynez, Shashi Narayan, Bernd Bohnet, and Ryan McDonald. 2020.
\newblock On faithfulness and factuality in abstractive summarization.
\newblock In \emph{Proceedings of the 58th Annual Meeting of the Association
  for Computational Linguistics}, pages 1906--1919.

\bibitem[{Nie et~al.(2019)Nie, Yao, Wang, Pan, and Lin}]{nie2019simple}
Feng Nie, Jin-Ge Yao, Jinpeng Wang, Rong Pan, and Chin-Yew Lin. 2019.
\newblock \href {https://doi.org/10.18653/v1/P19-1256} {A simple recipe towards
  reducing hallucination in neural surface realisation}.
\newblock In \emph{Proceedings of the 57th Annual Meeting of the Association
  for Computational Linguistics}, pages 2673--2679, Florence, Italy.
  Association for Computational Linguistics.

\bibitem[{Papineni et~al.(2002)Papineni, Roukos, Ward, and
  Zhu}]{Papineni2002BleuAM}
Kishore Papineni, Salim Roukos, Todd Ward, and Wei-Jing Zhu. 2002.
\newblock Bleu: a method for automatic evaluation of machine translation.
\newblock In \emph{ACL}.

\bibitem[{Paszke et~al.(2019)Paszke, Gross, Massa, Lerer, Bradbury, Chanan,
  Killeen, Lin, Gimelshein, Antiga, Desmaison, K{\"o}pf, Yang, DeVito, Raison,
  Tejani, Chilamkurthy, Steiner, Fang, Bai, and Chintala}]{Paszke2019PyTorchAI}
Adam Paszke, Sam Gross, Francisco Massa, Adam Lerer, James Bradbury, Gregory
  Chanan, Trevor Killeen, Zeming Lin, Natalia Gimelshein, Luca Antiga, Alban
  Desmaison, Andreas K{\"o}pf, Edward Yang, Zach DeVito, Martin Raison, Alykhan
  Tejani, Sasank Chilamkurthy, Benoit Steiner, Lu~Fang, Junjie Bai, and Soumith
  Chintala. 2019.
\newblock Pytorch: An imperative style, high-performance deep learning library.
\newblock In \emph{NeurIPS}.

\bibitem[{Pavlopoulos et~al.(2019)Pavlopoulos, Kougia, and
  Androutsopoulos}]{biomedical_ic}
John Pavlopoulos, Vasiliki Kougia, and Ion Androutsopoulos. 2019.
\newblock \href {https://doi.org/10.18653/v1/W19-1803} {A survey on biomedical
  image captioning}.
\newblock In \emph{Proceedings of the Second Workshop on Shortcomings in Vision
  and Language}, pages 26--36, Minneapolis, Minnesota. Association for
  Computational Linguistics.

\bibitem[{Radford et~al.(2021)Radford, Kim, Hallacy, Ramesh, Goh, Agarwal,
  Sastry, Askell, Mishkin, Clark, Krueger, and
  Sutskever}]{Radford2021LearningTV}
Alec Radford, Jong~Wook Kim, Chris Hallacy, Aditya Ramesh, Gabriel Goh,
  Sandhini Agarwal, Girish Sastry, Amanda Askell, Pamela Mishkin, Jack Clark,
  Gretchen Krueger, and Ilya Sutskever. 2021.
\newblock Learning transferable visual models from natural language
  supervision.
\newblock In \emph{ICML}.

\bibitem[{Ren et~al.(2015)Ren, He, Girshick, and Sun}]{Ren2015FasterRT}
Shaoqing Ren, Kaiming He, Ross~B. Girshick, and Jian Sun. 2015.
\newblock Faster r-cnn: Towards real-time object detection with region proposal
  networks.
\newblock \emph{IEEE Transactions on Pattern Analysis and Machine
  Intelligence}, 39:1137--1149.

\bibitem[{Rennie et~al.(2017)Rennie, Marcheret, Mroueh, Ross, and
  Goel}]{Rennie2017SelfCriticalST}
Steven~J. Rennie, Etienne Marcheret, Youssef Mroueh, Jerret Ross, and Vaibhava
  Goel. 2017.
\newblock Self-critical sequence training for image captioning.
\newblock \emph{2017 IEEE Conference on Computer Vision and Pattern Recognition
  (CVPR)}, pages 1179--1195.

\bibitem[{Robinson et~al.(2021)Robinson, Chuang, Sra, and
  Jegelka}]{Robinson2021ContrastiveLW}
Joshua~David Robinson, Ching-Yao Chuang, Suvrit Sra, and Stefanie Jegelka.
  2021.
\newblock \href {https://openreview.net/forum?id=CR1XOQ0UTh-} {Contrastive
  learning with hard negative samples}.
\newblock In \emph{International Conference on Learning Representations}.

\bibitem[{Rohrbach et~al.(2018)Rohrbach, Hendricks, Burns, Darrell, and
  Saenko}]{rohrbach2018object}
Anna Rohrbach, Lisa~Anne Hendricks, Kaylee Burns, Trevor Darrell, and Kate
  Saenko. 2018.
\newblock Object hallucination in image captioning.
\newblock In \emph{Proceedings of the 2018 Conference on Empirical Methods in
  Natural Language Processing}, pages 4035--4045.

\bibitem[{Sharma et~al.(2018)Sharma, Ding, Goodman, and
  Soricut}]{Sharma2018ConceptualCA}
Piyush Sharma, Nan Ding, Sebastian Goodman, and Radu Soricut. 2018.
\newblock Conceptual captions: A cleaned, hypernymed, image alt-text dataset
  for automatic image captioning.
\newblock In \emph{ACL}.

\bibitem[{Shen et~al.(2022)Shen, Li, Tan, Bansal, Rohrbach, Chang, Yao, and
  Keutzer}]{Shen2021HowMC}
Sheng Shen, Liunian~Harold Li, Hao Tan, Mohit Bansal, Anna Rohrbach, Kai-Wei
  Chang, Zhewei Yao, and Kurt Keutzer. 2022.
\newblock \href {https://openreview.net/forum?id=zf_Ll3HZWgy} {How much can
  {CLIP} benefit vision-and-language tasks?}
\newblock In \emph{International Conference on Learning Representations}.

\bibitem[{Shuster et~al.(2021)Shuster, Poff, Chen, Kiela, and
  Weston}]{shuster2021retrieval}
Kurt Shuster, Spencer Poff, Moya Chen, Douwe Kiela, and Jason Weston. 2021.
\newblock Retrieval augmentation reduces hallucination in conversation.
\newblock \emph{EMNLP}.

\bibitem[{Sigurdsson et~al.(2020)Sigurdsson, Alayrac, Nematzadeh, Smaira,
  Malinowski, Carreira, Blunsom, and Zisserman}]{Sigurdsson2020VisualGI}
Gunnar~A. Sigurdsson, Jean-Baptiste Alayrac, Aida Nematzadeh, Lucas Smaira,
  Mateusz Malinowski, Jo{\~a}o Carreira, Phil Blunsom, and Andrew Zisserman.
  2020.
\newblock Visual grounding in video for unsupervised word translation.
\newblock \emph{2020 IEEE/CVF Conference on Computer Vision and Pattern
  Recognition (CVPR)}, pages 10847--10856.

\bibitem[{Su et~al.(2022)Su, Li, Zhang, Shang, Jiang, Liu, and
  Fung}]{su2022read}
Dan Su, Xiaoguang Li, Jindi Zhang, Lifeng Shang, Xin Jiang, Qun Liu, and
  Pascale Fung. 2022.
\newblock Read before generate! faithful long form question answering with
  machine reading.
\newblock In \emph{Findings of the Association for Computational Linguistics:
  ACL 2022}, pages 744--756.

\bibitem[{Su et~al.(2020)Su, Zhu, Cao, Li, Lu, Wei, and Dai}]{Su2020VLBERTPO}
Weijie Su, Xizhou Zhu, Yue Cao, Bin Li, Lewei Lu, Furu Wei, and Jifeng Dai.
  2020.
\newblock \href {https://openreview.net/forum?id=SygXPaEYvH} {Vl-bert:
  Pre-training of generic visual-linguistic representations}.
\newblock In \emph{International Conference on Learning Representations}.

\bibitem[{Sun et~al.(2019)Sun, Myers, Vondrick, Murphy, and
  Schmid}]{Sun2019VideoBERTAJ}
Chen Sun, Austin Myers, Carl Vondrick, Kevin~P. Murphy, and Cordelia Schmid.
  2019.
\newblock Videobert: A joint model for video and language representation
  learning.
\newblock \emph{2019 IEEE/CVF International Conference on Computer Vision
  (ICCV)}, pages 7463--7472.

\bibitem[{Tan and Bansal(2019)}]{Tan2019LXMERTLC}
Hao Tan and Mohit Bansal. 2019.
\newblock \href {https://doi.org/10.18653/v1/D19-1514} {{LXMERT}: Learning
  cross-modality encoder representations from transformers}.
\newblock In \emph{Proceedings of the 2019 Conference on Empirical Methods in
  Natural Language Processing and the 9th International Joint Conference on
  Natural Language Processing (EMNLP-IJCNLP)}, pages 5100--5111, Hong Kong,
  China. Association for Computational Linguistics.

\bibitem[{Tian et~al.(2020)Tian, Krishnan, and Isola}]{Tian2020ContrastiveMC}
Yonglong Tian, Dilip Krishnan, and Phillip Isola. 2020.
\newblock Contrastive multiview coding.
\newblock In \emph{ECCV}.

\bibitem[{Vaswani et~al.(2017)Vaswani, Shazeer, Parmar, Uszkoreit, Jones,
  Gomez, Kaiser, and Polosukhin}]{Vaswani2017AttentionIA}
Ashish Vaswani, Noam Shazeer, Niki Parmar, Jakob Uszkoreit, Llion Jones,
  Aidan~N Gomez, \L~ukasz Kaiser, and Illia Polosukhin. 2017.
\newblock \href
  {https://proceedings.neurips.cc/paper/2017/file/3f5ee243547dee91fbd053c1c4a845aa-Paper.pdf}
  {Attention is all you need}.
\newblock In \emph{Advances in Neural Information Processing Systems},
  volume~30. Curran Associates, Inc.

\bibitem[{Vedantam et~al.(2015)Vedantam, Zitnick, and
  Parikh}]{Vedantam2015CIDErCI}
Ramakrishna Vedantam, C.~Lawrence Zitnick, and Devi Parikh. 2015.
\newblock Cider: Consensus-based image description evaluation.
\newblock \emph{2015 IEEE Conference on Computer Vision and Pattern Recognition
  (CVPR)}, pages 4566--4575.

\bibitem[{Wang et~al.(2022{\natexlab{a}})Wang, Yang, Men, Lin, Bai, Li, Ma,
  Zhou, Zhou, and Yang}]{Wang2022UnifyingAT}
Peng Wang, An~Yang, Rui Men, Junyang Lin, Shuai Bai, Zhikang Li, Jianxin Ma,
  Chang Zhou, Jingren Zhou, and Hongxia Yang. 2022{\natexlab{a}}.
\newblock \href {https://proceedings.mlr.press/v162/wang22al.html} {{OFA}:
  Unifying architectures, tasks, and modalities through a simple
  sequence-to-sequence learning framework}.
\newblock In \emph{Proceedings of the 39th International Conference on Machine
  Learning}, volume 162 of \emph{Proceedings of Machine Learning Research},
  pages 23318--23340. PMLR.

\bibitem[{Wang et~al.(2022{\natexlab{b}})Wang, Yu, Yu, Dai, Tsvetkov, and
  Cao}]{Wang2021SimVLMSV}
Zirui Wang, Jiahui Yu, Adams~Wei Yu, Zihang Dai, Yulia Tsvetkov, and Yuan Cao.
  2022{\natexlab{b}}.
\newblock \href {https://openreview.net/forum?id=GUrhfTuf_3} {Sim{VLM}: Simple
  visual language model pretraining with weak supervision}.
\newblock In \emph{International Conference on Learning Representations}.

\bibitem[{Xiao and Wang(2021)}]{xiao2021hallucination}
Yijun Xiao and William~Yang Wang. 2021.
\newblock On hallucination and predictive uncertainty in conditional language
  generation.
\newblock In \emph{Proceedings of the 16th Conference of the European Chapter
  of the Association for Computational Linguistics: Main Volume}, pages
  2734--2744.

\bibitem[{Xie et~al.(2017)Xie, Girshick, Doll{\'a}r, Tu, and
  He}]{Xie2017AggregatedRT}
Saining Xie, Ross~B. Girshick, Piotr Doll{\'a}r, Zhuowen Tu, and Kaiming He.
  2017.
\newblock Aggregated residual transformations for deep neural networks.
\newblock \emph{2017 IEEE Conference on Computer Vision and Pattern Recognition
  (CVPR)}, pages 5987--5995.

\bibitem[{Yao et~al.(2022)Yao, Huang, Hou, Lu, Niu, Xu, Liang, Li, Jiang, and
  Xu}]{Yao2021FILIPFI}
Lewei Yao, Runhui Huang, Lu~Hou, Guansong Lu, Minzhe Niu, Hang Xu, Xiaodan
  Liang, Zhenguo Li, Xin Jiang, and Chunjing Xu. 2022.
\newblock \href {https://openreview.net/forum?id=cpDhcsEDC2} {{FILIP}:
  Fine-grained interactive language-image pre-training}.
\newblock In \emph{International Conference on Learning Representations}.

\bibitem[{Yu et~al.(2021{\natexlab{a}})Yu, Tang, Yin, Sun, Tian, Wu, and
  Wang}]{Yu_Tang_Yin_Sun_Tian_Wu_Wang_2021}
Fei Yu, Jiji Tang, Weichong Yin, Yu~Sun, Hao Tian, Hua Wu, and Haifeng Wang.
  2021{\natexlab{a}}.
\newblock \href {https://doi.org/10.1609/aaai.v35i4.16431} {Ernie-vil:
  Knowledge enhanced vision-language representations through scene graphs}.
\newblock \emph{Proceedings of the AAAI Conference on Artificial Intelligence},
  35(4):3208--3216.

\bibitem[{Yu et~al.(2021{\natexlab{b}})Yu, Dai, Liu, and
  Fung}]{yu-etal-2021-vision}
Tiezheng Yu, Wenliang Dai, Zihan Liu, and Pascale Fung. 2021{\natexlab{b}}.
\newblock \href {https://doi.org/10.18653/v1/2021.emnlp-main.326} {Vision
  guided generative pre-trained language models for multimodal abstractive
  summarization}.
\newblock In \emph{Proceedings of the 2021 Conference on Empirical Methods in
  Natural Language Processing}, pages 3995--4007, Online and Punta Cana,
  Dominican Republic. Association for Computational Linguistics.

\bibitem[{Zhai et~al.(2022)Zhai, Wang, Mustafa, Steiner, Keysers, Kolesnikov,
  and Beyer}]{Zhai2021LiTZT}
Xiaohua Zhai, Xiao Wang, Basil Mustafa, Andreas Steiner, Daniel Keysers,
  Alexander Kolesnikov, and Lucas Beyer. 2022.
\newblock Lit: Zero-shot transfer with locked-image text tuning.
\newblock In \emph{Proceedings of the IEEE/CVF Conference on Computer Vision
  and Pattern Recognition (CVPR)}, pages 18123--18133.

\bibitem[{Zhang et~al.(2021{\natexlab{a}})Zhang, Li, Hu, Yang, Zhang, Wang,
  Choi, and Gao}]{Zhang2021VinVLRV}
Pengchuan Zhang, Xiujun Li, Xiaowei Hu, Jianwei Yang, Lei Zhang, Lijuan Wang,
  Yejin Choi, and Jianfeng Gao. 2021{\natexlab{a}}.
\newblock Vinvl: Revisiting visual representations in vision-language models.
\newblock \emph{2021 IEEE/CVF Conference on Computer Vision and Pattern
  Recognition (CVPR)}, pages 5575--5584.

\bibitem[{Zhang et~al.(2021{\natexlab{b}})Zhang, Shi, Tang, Xiao, Yu, and
  Zhuang}]{Zhang_Shi_Tang_Xiao_Yu_Zhuang_2021}
Wenqiao Zhang, Haochen Shi, Siliang Tang, Jun Xiao, Qiang Yu, and Yueting
  Zhuang. 2021{\natexlab{b}}.
\newblock \href {https://doi.org/10.1609/aaai.v35i4.16452} {Consensus graph
  representation learning for better grounded image captioning}.
\newblock \emph{Proceedings of the AAAI Conference on Artificial Intelligence},
  35(4):3394--3402.

\bibitem[{Zhang et~al.(2020)Zhang, Wang, Tang, Shi, Shi, Xiao, Zhuang, and
  Wang}]{10.1145/3394171.3413746}
Wenqiao Zhang, Xin~Eric Wang, Siliang Tang, Haizhou Shi, Haochen Shi, Jun Xiao,
  Yueting Zhuang, and William~Yang Wang. 2020.
\newblock \href {https://doi.org/10.1145/3394171.3413746} {Relational graph
  learning for grounded video description generation}.
\newblock In \emph{Proceedings of the 28th ACM International Conference on
  Multimedia}, MM '20, page 3807–3828, New York, NY, USA. Association for
  Computing Machinery.

\bibitem[{Zhou et~al.(2020)Zhou, Palangi, Zhang, Hu, Corso, and
  Gao}]{Zhou2020UnifiedVP}
Luowei Zhou, Hamid Palangi, Lei Zhang, Houdong Hu, Jason~J. Corso, and Jianfeng
  Gao. 2020.
\newblock Unified vision-language pre-training for image captioning and vqa.
\newblock \emph{ArXiv}, abs/1909.11059.

\bibitem[{Zhou et~al.(2021)Zhou, Zhou, Wang, Cheng, Li, Yu, and
  Liu}]{Zhou2021UC2UC}
Mingyang Zhou, Luowei Zhou, Shuohang Wang, Yu~Cheng, Linjie Li, Zhou Yu, and
  Jingjing Liu. 2021.
\newblock Uc2: Universal cross-lingual cross-modal vision-and-language
  pre-training.
\newblock \emph{2021 IEEE/CVF Conference on Computer Vision and Pattern
  Recognition (CVPR)}, pages 4153--4163.

\end{thebibliography}
\bibliographystyle{acl_natbib}

\newpage
\onecolumn

\appendix

\section{Implementation Details}
\label{appendix:implementation_details}
Our experiments are implemented in the PyTorch framework~\citep{Paszke2019PyTorchAI}. For both pre-training and finetuning, we use 8 Nvidia V100 GPUs. As mentioned in Section~\ref{sec:model_architecture}, we use the official CLIP checkpoints provided on GitHub. For the text decoder BertLM, we initialize model weights from the bert-base-uncased checkpoint with 110M parameters. For the finetuning on COCO Caption, we use a batch size of 512 and train the models with the AdamW optimizer~\citep{Loshchilov2019DecoupledWD} for 10 epochs with a learning rate of $5 \times 10^{-5}$ and a weight decay of $1 \times 10^{-2}$. The learning rate is decayed linearly after each epoch with a rate of 0.85. For the pre-training of text generation losses (LM and ObjMLM), we keep the same hyper-parameters with a learning rate warmup within the first epoch. For ITC and ITM losses, we increase the batch size to 1024 as they tend to have a better performance with more negative samples.

\section{Additional Case Studies}
\label{appendix:additional_cases}

\begin{figure*}[h!]
    \centering
    \includegraphics[width=0.87\linewidth]{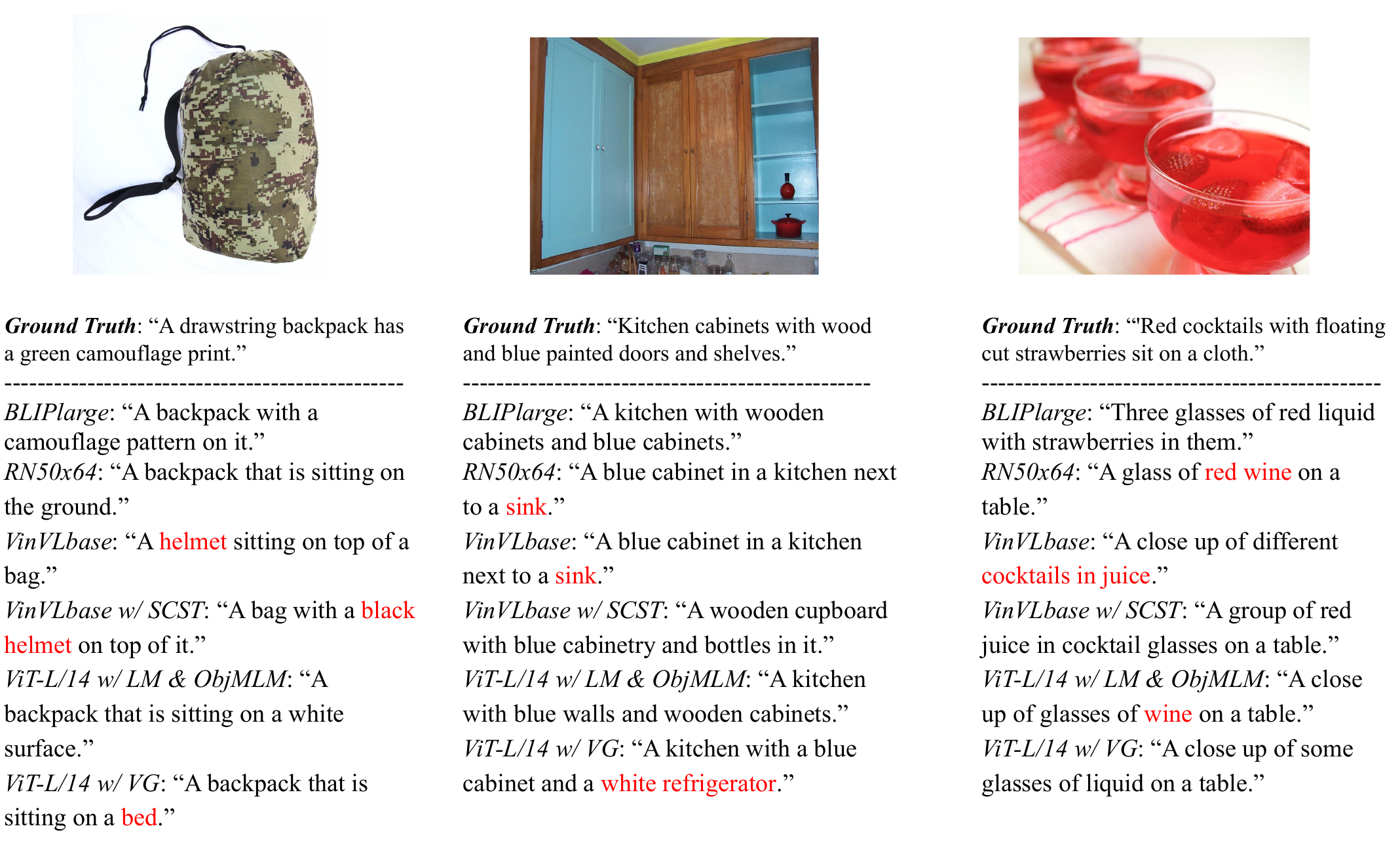}
    \label{fig:appendix_example-1}
\end{figure*}
\vspace{-20pt}
\begin{figure*}[!h]
    \centering
    \includegraphics[width=0.87\linewidth]{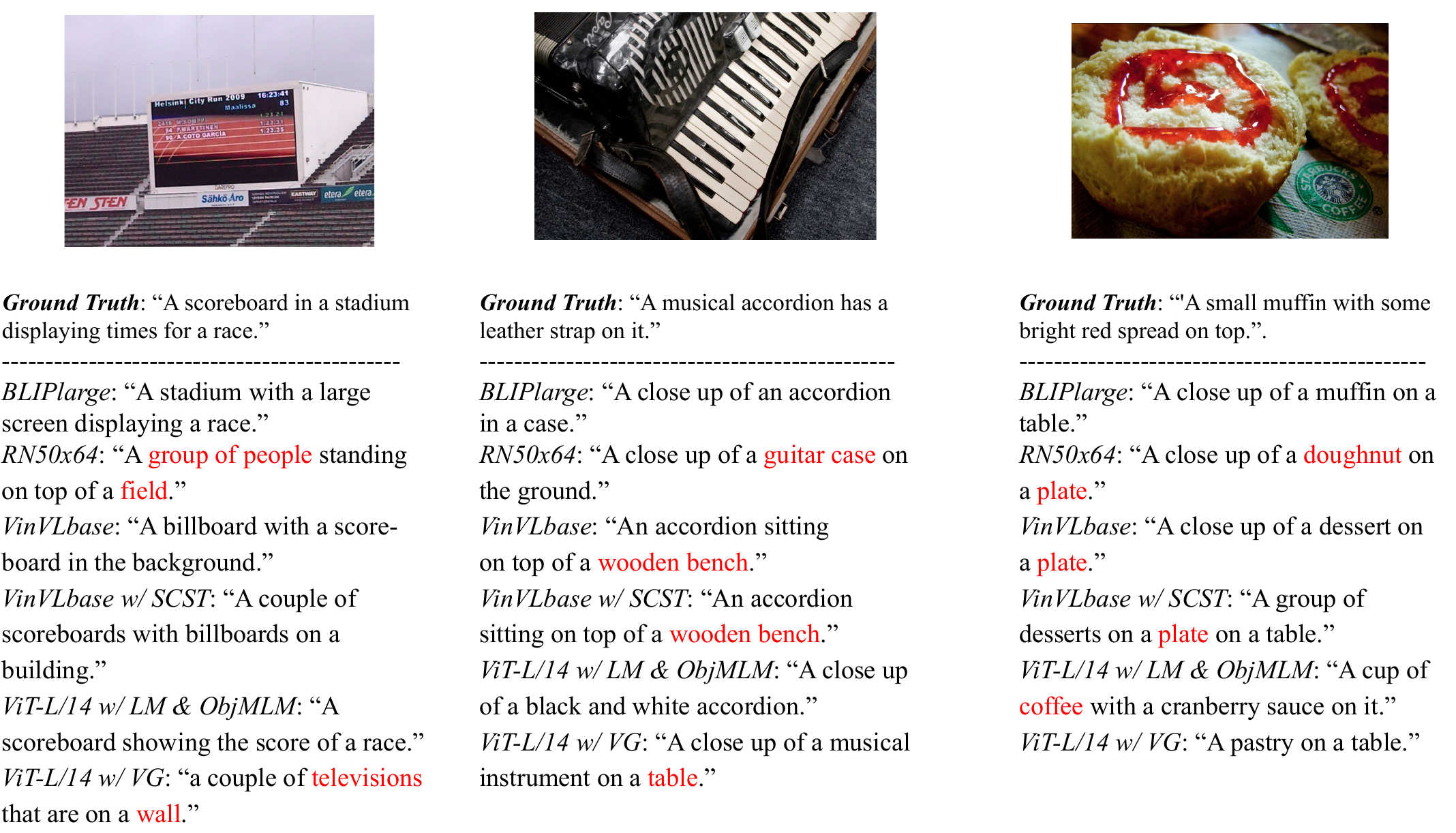}
    \caption{More cases of generated captions from different models, where the hallucinated objects are marked in red.}
    \label{fig:appendix_example-2}
\end{figure*}

\end{document}